\documentclass[letterpaper, 10pt, conference]{ieeeconf}

\IEEEoverridecommandlockouts
\usepackage[utf8]{inputenc}
\usepackage{amsmath,amssymb,amsfonts}
\usepackage{algorithmic}
\usepackage{algorithm}
\usepackage{graphicx}
\usepackage{textcomp}
\usepackage{xcolor}
\usepackage{booktabs}
\usepackage{multirow}
\usepackage{hyperref}
\hypersetup{colorlinks=true,linkcolor=black,citecolor=black,urlcolor=blue}
\usepackage{cite}
\usepackage{subcaption}
\usepackage{tabularx}
\let\labelindent\relax 
\usepackage{enumitem}
\usepackage{url}
\usepackage{times}
\usepackage{tikz}
\usepackage{pgfplots}
\pgfplotsset{compat=1.18}
\usetikzlibrary{
  positioning, arrows.meta, shapes.geometric, shapes.multipart,
  fit, backgrounds, calc, decorations.pathreplacing, patterns,
  shadows.blur, chains
}

\definecolor{inputblue}{HTML}{D6EAF8}
\definecolor{inputborder}{HTML}{2E86C1}
\definecolor{plannerorg}{HTML}{FDEBD0}
\definecolor{plannerborder}{HTML}{E67E22}
\definecolor{memgreen}{HTML}{D5F5E3}
\definecolor{memborder}{HTML}{27AE60}
\definecolor{execpink}{HTML}{FADBD8}
\definecolor{execborder}{HTML}{E74C3C}
\definecolor{skillblue}{HTML}{D4E6F1}
\definecolor{skillborder}{HTML}{2980B9}
\definecolor{arrowred}{HTML}{C0392B}
\definecolor{arrowgreen}{HTML}{27AE60}
\definecolor{arrowblue}{HTML}{2E86C1}
\definecolor{arrowgray}{HTML}{7F8C8D}
\definecolor{reflectpurple}{HTML}{8E44AD}
\definecolor{successteal}{HTML}{17A589}
\definecolor{textdarkgray}{HTML}{5D6D7E}
\definecolor{l1gold}{HTML}{F39C12}
\definecolor{l2green}{HTML}{27AE60}
\definecolor{l3purple}{HTML}{8E44AD}
\definecolor{fragred}{HTML}{CB4335}
\definecolor{kborange}{HTML}{E67E22}
\definecolor{constpurple}{HTML}{8E44AD}
\definecolor{adjustteal}{HTML}{17A589}
\definecolor{sgblue}{HTML}{2E86C1}
\definecolor{itgreen}{HTML}{27AE60}
\definecolor{lhorange}{HTML}{E67E22}
\definecolor{framebg}{HTML}{F8F9FA}
\definecolor{lvl1blue}{HTML}{3498DB}
\definecolor{lvl2org}{HTML}{E67E22}
\definecolor{lvl3red}{HTML}{C0392B}

\setlist{noitemsep,topsep=1pt,parsep=1pt,partopsep=0pt}

\AtBeginDocument{%
  \abovedisplayskip=4pt plus 2pt minus 2pt
  \belowdisplayskip=4pt plus 2pt minus 2pt
  \abovedisplayshortskip=2pt plus 1pt
  \belowdisplayshortskip=2pt plus 1pt
}

\newcommand{\sysname}{Agentic RAG-VLM}
\newcommand{\haarag}{HAA-RAG}

\def\BibTeX{{\rm B\kern-.05em{\sc i\kern-.025em b}\kern-.08em
    T\kern-.1667em\lower.7ex\hbox{E}\kern-.125emX}}

\begin{document}

\title{\LARGE \bf Agentic RAG-VLM: Affordance-Aware Retrieval-Augmented\\Generation with Self-Reflective Planning for Robotic Grasping}

\author{Tao Chen$^{1}$, Lizheng Liu$^{1\dagger}$, Jiaxu Wang$^{1}$, Ziyue Jiang$^{1}$, Ruiqi Tian$^{2}$, JiGuang Huo$^{1}$, Zhongxue Gan$^{1}$%
\thanks{$^{1}$Tao Chen, Lizheng Liu, Jiaxu Wang, Ziyue Jiang, JiGuang Huo, and Zhongxue Gan are with Fudan University, Shanghai, China (e-mail: {\tt\small lzliu@fudan.edu.cn}).}%
\thanks{$^{2}$Ruiqi Tian is with Kean University.}%
\thanks{$^{\dagger}$Corresponding author.}%
}

\maketitle
\thispagestyle{empty}
\pagestyle{empty}

\begin{abstract}
Generalizable robotic grasping in cluttered environments is essential for deploying manipulators in unstructured human spaces, yet existing VLM-based methods rely on visual similarity for object matching---neglecting physical affordances such as handle graspability and material fragility---and operate open-loop without spatial reasoning or failure recovery, limiting their effectiveness when objects are densely packed or physically diverse.
We present \sysname{}, a unified framework that bridges VLM-based semantic understanding and physically grounded grasp execution by integrating retrieval-augmented generation~(RAG) with vision-language models~(VLMs) and agentic self-reflective planning.
\sysname{} introduces three tightly coupled components:
(1)~a Hierarchical Affordance-Aware RAG (\haarag{}) that encodes four-dimensional affordance descriptors---type, material, fragility, and graspable region---and retrieves strategies by functional affordance compatibility rather than visual appearance;
(2)~a Scene Graph Constraint Reasoner that constructs spatial relationship graphs from VLM perception and translates proximity, occlusion, and support constraints into concrete grasp parameter adjustments;
and (3)~an Agentic Self-Reflective Pipeline with a 14-type failure taxonomy and three-level adaptive retry for closed-loop grasp refinement.
Evaluated on a 12-task benchmark spanning single-grasp, interactive, and long-horizon scenarios with 360 trials per configuration, \sysname{} achieves 78.3\% overall success---a 53.3\,pp absolute gain over VLM-only baselines---demonstrating that affordance-aware retrieval, scene graph reasoning, and agentic recovery are jointly essential for robust manipulation.
\end{abstract}

\section{Introduction}
\label{sec:intro}

Robotic grasping in unstructured environments requires integrating visual perception, semantic understanding, and physical execution.
A general-purpose grasping pipeline typically involves perceiving the scene from sensor inputs, planning a grasp strategy, and executing it through a manipulator.
While each stage has seen significant progress independently---from learning-based grasp detection~\cite{kumra2020grconvnet, fang2023anygrasp} to language-conditioned imitation~\cite{shridhar2022cliport}---integrating them into a coherent system that handles diverse objects in cluttered, multi-object environments remains largely unsolved.
This is especially hard for long-horizon tasks like table clearing, where sequential dependencies and accumulated errors require both accurate planning and adaptive failure recovery.

Recent vision-language models (VLMs) have opened new directions for bridging perception and planning.
RT-2~\cite{brohan2023rt2} pioneers end-to-end VLM-to-action by tokenizing robotic actions alongside language; SayCan~\cite{ahn2022saycan} decouples high-level planning from low-level execution through learned affordance functions; VoxPoser~\cite{huang2023voxposer} and Code as Policies~\cite{liang2023code} generate 3D value maps and executable programs, respectively; and ManipLLM~\cite{li2024manipllm} fine-tunes multimodal LLMs for 6-DoF grasp prediction.
These advances point toward general-purpose language-conditioned manipulation.
Yet three gaps remain:

\textbf{Semantic-Manipulation Gap.}
VLM-based systems rely on visual similarity for object matching, but visual resemblance does not imply manipulation compatibility.
A ceramic mug and a glass vase may occupy nearby regions in CLIP~\cite{radford2021clip} embedding space---both are cylindrical, hollow, and similarly sized---yet the mug should be grasped by its handle with a firm grip, while the vase requires a cautious side approach with reduced force.
This mismatch between visual similarity and manipulation compatibility~\cite{gibson1977affordance} is largely unaddressed.

\textbf{Scene-Unaware Planning.}
Most systems reason about the target object in isolation.
In cluttered environments, a cup placed 5\,cm from a fragile wine glass necessitates a cautious lateral approach with reduced force, whereas the same cup on an empty table can be grasped directly from above.
Scene graphs have been adopted for task and motion planning~\cite{zeng2020transporter}, but typically demand ground-truth annotations and do not translate spatial reasoning into concrete grasp parameter adjustments.
SpatialVLM~\cite{kwon2024spatialvlm} endows VLMs with spatial understanding but does not connect it to manipulation constraints.

\textbf{Failure Recovery.}
Classical grasp planners~\cite{kumra2020grconvnet, fang2023anygrasp, sundermeyer2021contact} operate open-loop, generating grasp poses without learning from failures.
Inner Monologue~\cite{huang2022inner} introduces VLM-based feedback but provides only unstructured observations (``the object slipped'') without systematic classification.
Reflexion~\cite{shinn2023reflexion} demonstrates that verbal self-reflection improves sequential decision-making, yet applying reflection to physical manipulation requires translating linguistic feedback into concrete, physics-grounded corrections---``the object slipped'' must map to ``increase grip force by 20\%,'' not merely a rephrased retry.

\begin{figure*}[t]
    \centering
    \includegraphics[width=\textwidth]{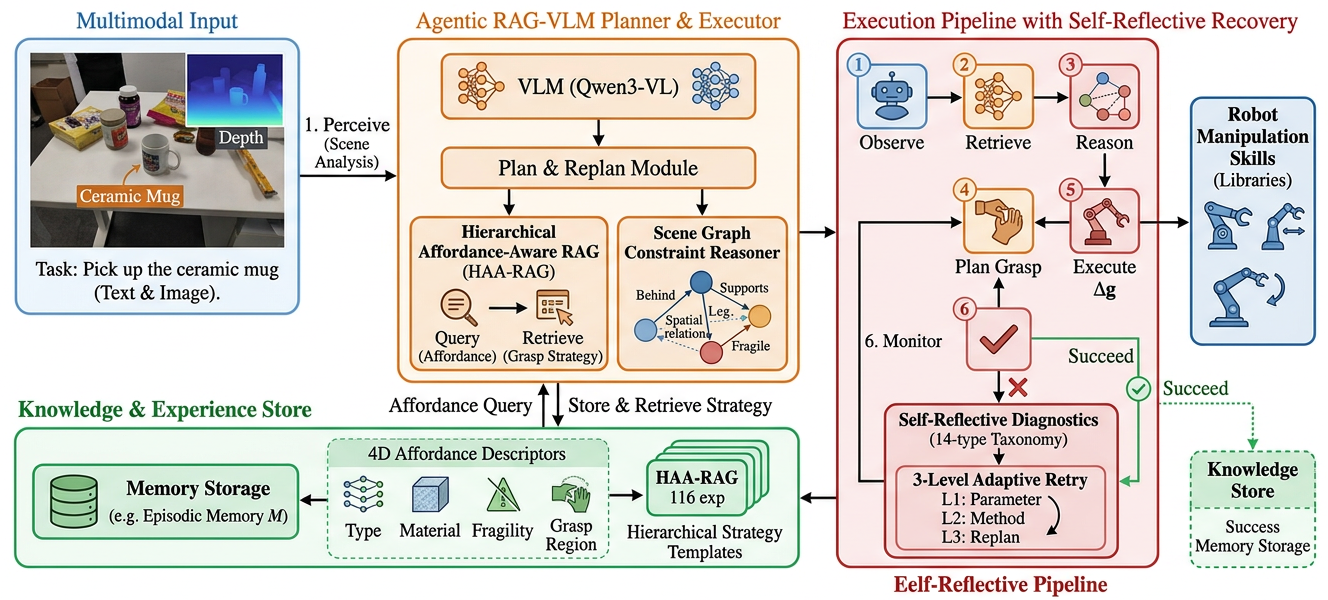}
    \caption{Overview of \sysname{}. The VLM-based Task Planner processes multimodal input and orchestrates a seven-stage execution pipeline: scene perception, affordance-aware retrieval via HAA-RAG, scene graph constraint reasoning ($\Delta\mathbf{g}$), and grasp execution. Upon failure ($\times$), the recovery loop diagnoses failures via a 14-type taxonomy, escalates corrections through three levels (L1: parameter tuning $\to$ L2: method switch $\to$ L3: full replan), and stores successful grasps in episodic memory $\mathcal{M}$ for cross-task transfer.}
    \label{fig:system}
\end{figure*}

To address these gaps, we propose \sysname{} (Fig.~\ref{fig:system}), a unified framework integrating affordance-aware experience retrieval, scene graph constraint reasoning, and agentic self-reflective planning for robust robotic grasping.
The contributions are summarized as follows:
\begin{itemize}
    \item We propose \sysname{}, a unified framework that integrates affordance-aware retrieval-augmented generation with scene graph constraint reasoning and agentic self-reflective planning, bridging the gap between VLM-based semantic understanding and physically grounded grasp execution.
    
    \item The framework introduces Hierarchical Affordance-Aware RAG (\haarag{}), a three-level retrieval pipeline that matches grasp experiences by functional affordance rather than visual similarity, and a Scene Graph Constraint Reasoner that translates spatial relationships into concrete grasp parameter adjustments for safe manipulation in cluttered environments.
    
    \item An Agentic Self-Reflective Pipeline that employs structured failure diagnosis with a 14-type taxonomy and three-level adaptive retry, enabling the system to learn from failed attempts and progressively refine grasp execution through physics-grounded corrections.
\end{itemize}

\section{Related Work}
\label{sec:related}

\textbf{VLMs for Manipulation.}
Recent VLM-based approaches range from end-to-end action tokenization (RT-2~\cite{brohan2023rt2}) to modular planning with affordance grounding (SayCan~\cite{ahn2022saycan}), 3D value maps (VoxPoser~\cite{huang2023voxposer}), code generation (Code as Policies~\cite{liang2023code}), and fine-tuned grasp prediction (ManipLLM~\cite{li2024manipllm}, RoboDexVLM~\cite{liu2025robodexvlm}, GPT-4V~\cite{wake2023gpt4v}).
However, these rely on parametric knowledge alone without retrieved experience or structured failure recovery.

\textbf{RAG and Affordance in Robotics.}
RAG~\cite{lewis2020rag} reduces hallucination by grounding generation in retrieved documents, but its robotic application remains nascent.
REFLECT~\cite{liu2023reflect} summarizes failure experiences without integrating retrieval into planning.
Meanwhile, affordance reasoning~\cite{gibson1977affordance} has been realized through pixel-level maps~\cite{mandikal2021dexvip} and part-based analysis~\cite{mo2021where2act}, but these predict \emph{where} to grasp, not \emph{how}---they identify graspable regions on an object surface without specifying the complete grasp strategy (force, width, approach direction, grasp type) appropriate for that object's physical affordance.
Our \haarag{} bridges this gap by explicitly encoding four-dimensional affordance descriptors (type, material, fragility, graspable region) and retrieving complete, physically grounded grasp strategies matched by functional affordance compatibility rather than visual similarity.

\textbf{Failure Recovery and Scene Reasoning.}
Classical planners~\cite{kumra2020grconvnet, fang2023anygrasp, sundermeyer2021contact} operate open-loop.
Inner Monologue~\cite{huang2022inner} and Reflexion~\cite{shinn2023reflexion} introduce verbal feedback but lack physics-grounded corrections for manipulation.
Scene graphs have been used for task planning~\cite{zeng2020transporter} but typically require ground-truth annotations.
SpatialVLM~\cite{kwon2024spatialvlm} adds spatial reasoning to VLMs without connecting it to grasp constraints.
Our pipeline addresses both gaps with a 14-type failure taxonomy mapping to quantitative corrections and VLM-constructed scene graphs that infer manipulation constraints without annotations.
In contrast to these approaches, \sysname{} uniquely integrates all five capabilities---affordance-aware retrieval, RAG-based experience grounding, scene graph constraint reasoning, structured failure recovery, and multi-step agentic planning---within a single unified framework.

\section{Affordance-Aware Retrieval and\texorpdfstring{\\}{ }Constraint Reasoning}
\label{sec:retrieval}

This section details how the system transforms a scene observation into a physically grounded grasp plan through two complementary stages: affordance-aware experience retrieval (Sec.~\ref{sec:haarag}) and spatial constraint reasoning (Sec.~\ref{sec:scene_graph}).
Given an RGB-D observation $I$ and a natural language instruction $q$, the system produces a grasp action $\mathbf{g} = (\mathbf{p}, \mathbf{d}, w, f, \tau)$---position, approach direction, gripper aperture, grip force, and grasp type ($\tau \in \{$power, pinch, side$\}$)---refined through three modular stages: (1)~affordance-aware retrieval via \haarag{}, (2)~scene graph constraint reasoning, and (3)~agentic planning with self-reflective retry (Sec.~\ref{sec:agentic}).

\subsection{Hierarchical Affordance-Aware RAG}
\label{sec:haarag}

\begin{figure}[t]
    \centering
    \includegraphics[width=\linewidth]{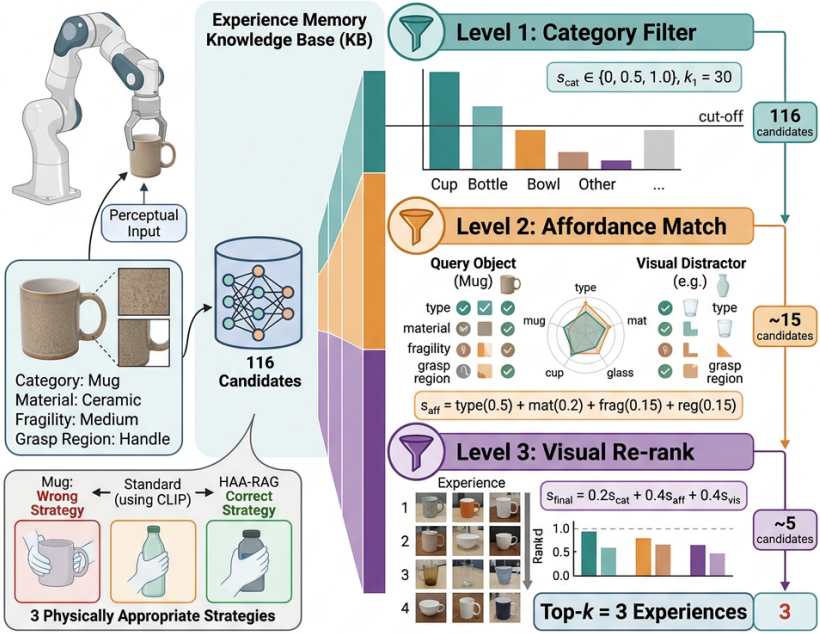}
    \caption{Hierarchical Affordance-Aware RAG (\haarag{}) pipeline. Experiences are progressively filtered through three levels: category matching, affordance scoring, and visual re-ranking, reducing 116 candidates to 3 physically appropriate strategies. Right annotations show candidate count at each stage.}
    \label{fig:haarag}
\end{figure}

Standard RAG retrieves by visual similarity (e.g., CLIP~\cite{radford2021clip} embeddings), but a bowl must be grasped along its rim while a cup uses its handle---visual similarity yields wrong strategies.
This is precisely the ``neglecting object-specific physical affordances'' problem identified in Sec.~\ref{sec:intro}: visually similar objects (e.g., a ceramic mug and a glass vase) may require fundamentally different manipulation strategies due to their distinct affordance properties.
\haarag{} directly addresses this by matching experiences through \emph{affordance compatibility} instead of visual resemblance (Fig.~\ref{fig:haarag}).

The foundation of this approach is a curated knowledge base $\mathcal{K} = \{e_1, \dots, e_N\}$ with $N = 116$ manipulation experiences, where each entry $e_i = (c_i, \mathbf{a}_i, \mathbf{g}_i, y_i, \mathbf{v}_i)$ stores:
\begin{itemize}
    \item $c_i$: Object category (e.g., ``mug'', ``bowl'', ``screwdriver'')
    \item $\mathbf{a}_i = (a^{\mathrm{type}}_i, a^{\mathrm{mat}}_i, a^{\mathrm{frag}}_i, a^{\mathrm{reg}}_i)$: Affordance descriptor with primary type (8 categories: \textsc{Graspable\_Body\slash Handle\slash Edge}, \textsc{Pinchable}, \textsc{Wrappable}, \textsc{Fragile}, \textsc{Tool\_Handle}, \textsc{Clampable}), material, fragility $a^{\mathrm{frag}} \in [0,1]$, and graspable region
    \item $\mathbf{g}_i$: Grasp parameters that led to the outcome
    \item $y_i \in \{0, 1\}$: Binary success label
    \item $\mathbf{v}_i$: Visual feature embedding (CLIP ViT-L/14)
\end{itemize}

Failed experiences ($y_i = 0$) are also retained with a contrastive penalty (scores reduced by 30\%) to discourage repeating failed strategies while remaining available as negative examples for the reflection module (Sec.~\ref{sec:failure}).

Leveraging this knowledge base, \haarag{} implements a three-level coarse-to-fine retrieval pipeline.
In the first level, the VLM identifies the target object category $c_q$ from instruction $q$ and the RGB image, retrieving the top-$k_1 = 30$ candidates from $\mathcal{K}$ by category match: exact matches score $s_{\mathrm{cat}} = 1.0$, same-superclass fuzzy matches (e.g., ``mug'' $\to$ ``cup'') score $0.5$, and unrelated categories are discarded.

In the second level, surviving candidates are re-scored by affordance similarity, capturing functional compatibility between the query object and stored experiences:
\begin{equation}
\begin{split}
    s_{\mathrm{aff}}(\mathbf{a}_q, \mathbf{a}_i) &= \underbrace{0.5 \cdot \mathbb{1}[a_q^{\mathrm{type}} = a_i^{\mathrm{type}}]}_{\text{affordance type match}} + \underbrace{0.2 \cdot s_{\mathrm{mat}}}_{\text{material}} \\
    &\quad + \underbrace{0.15 \cdot (1 - |\Delta f|)}_{\text{fragility}} + \underbrace{0.15 \cdot s_{\mathrm{reg}}}_{\text{region}}
\end{split}
    \label{eq:affordance_sim}
\end{equation}
where $s_{\mathrm{mat}} = \mathbb{1}[a_q^{\mathrm{mat}} = a_i^{\mathrm{mat}}]$ is material compatibility (e.g., both ceramic), $\Delta f = a_q^{\mathrm{frag}} - a_i^{\mathrm{frag}}$ captures fragility similarity, and $s_{\mathrm{reg}} = \mathbb{1}[a_q^{\mathrm{reg}} = a_i^{\mathrm{reg}}]$ is graspable region overlap.
The dominant weight on affordance type ($0.5$) ensures that functional compatibility is the primary selection criterion.
Candidates scoring below threshold $\tau_{\mathrm{aff}} = 0.3$ are discarded.

Finally, the remaining candidates are re-ranked by CLIP visual similarity $s_{\mathrm{vis}} = \mathrm{cos}(\mathbf{v}_q, \mathbf{v}_i)$ using the current RGB observation's embedding, ensuring that among affordance-compatible experiences, the most visually similar one is selected for fine-grained parameter matching.
The final retrieval score fuses all three levels (Eq.~\ref{eq:affordance_sim}) as $s_{\mathrm{final}} = 0.2\, s_{\mathrm{cat}} + 0.4\, s_{\mathrm{aff}} + 0.4\, s_{\mathrm{vis}}$, and the top-$k = 3$ experiences are returned to the planner as candidate strategies.

\subsection{Scene Graph Constraint Reasoning}
\label{sec:scene_graph}

While \haarag{} determines \emph{how} to grasp the target object based on its intrinsic affordance properties, real-world manipulation must also account for the \emph{extrinsic spatial context} imposed by neighboring objects.
The Scene Graph Constraint Reasoner addresses this by constructing a structured spatial representation and translating it into concrete grasp parameter adjustments.

From the VLM's scene analysis, we construct a directed graph $\mathcal{G} = (\mathcal{V}, \mathcal{E})$ where each node $v_j \in \mathcal{V}$ encodes object attributes $(\mathbf{p}_j, c_j, \mathrm{state}_j, a_j^{\mathrm{frag}})$---position, category, physical state (empty, filled, or stacked), and fragility score---and each directed edge $e_{jk} = (v_j, v_k, r_{jk}) \in \mathcal{E}$ captures a spatial relation $r_{jk}$ from 16 predefined types including \textsc{On\_Top\_Of}, \textsc{Contains}, \textsc{Occludes}, \textsc{Supports}, \textsc{Adjacent\_To}, and \textsc{Behind}.
A rule-based Constraint Analyzer then traverses $\mathcal{G}$ starting from the target node $v_t$, examining $v_t$'s attributes and neighbors within a radius of $\delta_{\mathrm{check}} = 0.15$\,m to infer active constraints $\mathcal{C}_{v_t}$.
Four constraint types are defined, each mapping to specific parameter adjustments:

\begin{enumerate}
    \item \textit{Content Preservation} ($\mathcal{C}_{\mathrm{content}}$): For filled containers---constrains approach to vertical ($\mathbf{d} = [0, 0, -1]$) with slow velocity to prevent spillage.
    
    \item \textit{Collision Avoidance} ($\mathcal{C}_{\mathrm{collision}}$): When fragile neighbor $v_f$ within $\delta_{\mathrm{safe}} = 0.10$\,m---applies force reduction $f' = 0.8f$, increases approach height by 30\,mm, and biases approach direction away from $v_f$.
    
    \item \textit{Support Dependency} ($\mathcal{C}_{\mathrm{support}}$): When a \textsc{Supports} edge $(v_t, v_s)$ exists, the system inserts a prerequisite action to remove the supported object~$v_s$ before attempting to grasp~$v_t$.
    
    \item \textit{Occlusion Handling} ($\mathcal{C}_{\mathrm{occlusion}}$): When occluder blocks direct line-of-sight to target---modifies approach to lateral path that avoids the occluding obstacle.
\end{enumerate}

\begin{figure}[b]
    \centering
    \includegraphics[width=\linewidth]{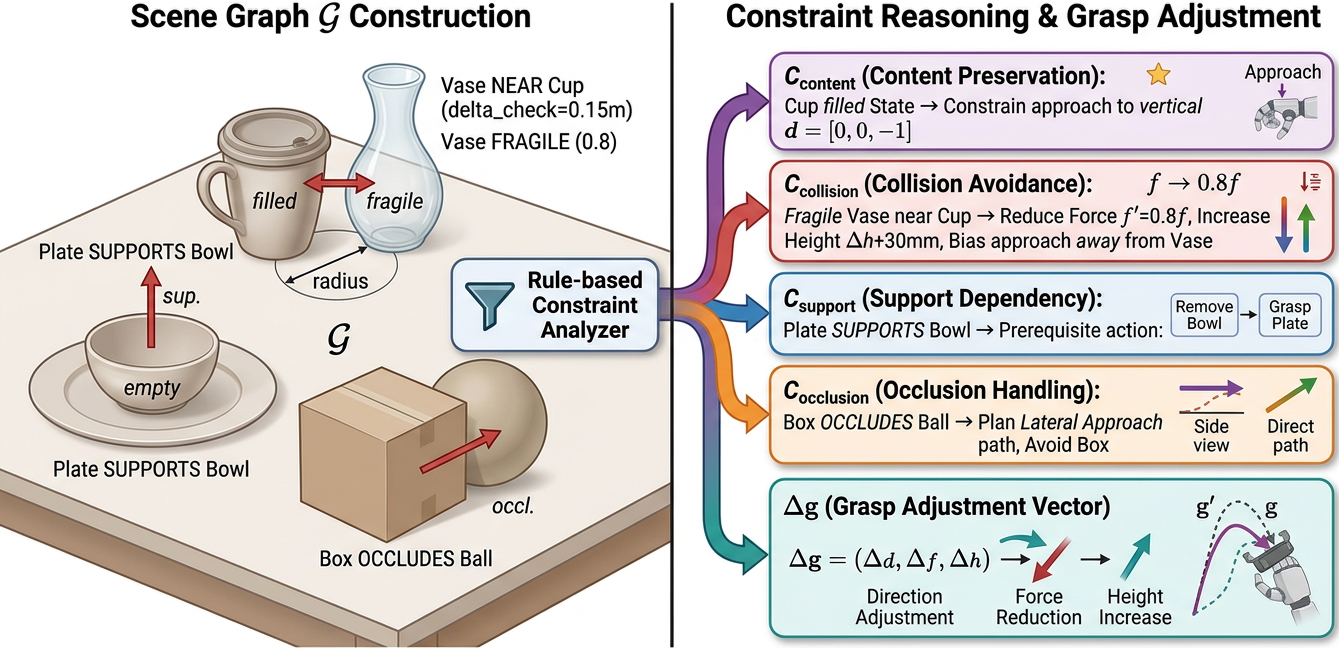}
    \caption{Scene graph constraint reasoning. Left: scene graph $\mathcal{G}$ with object nodes and spatial relations. Right: four constraint types inferred from the graph, each mapped to concrete parameter adjustments $\Delta\mathbf{g}$ that modify the planned grasp.}
    \label{fig:scene_graph}
\end{figure}

The inferred constraints are compiled into a \textbf{constraint adjustment vector} $\Delta\mathbf{g} = (\Delta\mathbf{d}, \Delta f, \Delta h)$ that additively modifies the planned grasp action: $\mathbf{g}' = \mathbf{g} \oplus \Delta\mathbf{g}$.
Multiple active constraints are composed sequentially, with collision avoidance taking priority (Fig.~\ref{fig:scene_graph}).

\section{Agentic Self-Reflective Grasp Planning}
\label{sec:agentic}

Real-world execution is noisy, and initial failures are common.
\sysname{} employs an Agentic Self-Reflective Pipeline providing \emph{closed-loop} execution with structured failure diagnosis and adaptive recovery.

\subsection{ReAct-Style Planning Loop}
\label{sec:react}

Inspired by the ReAct framework~\cite{yao2023react}, the VLM iterates through structured \textsc{Thought}$\rightarrow$\textsc{Action}$\rightarrow$\textsc{Observation} cycles until either a successful grasp is achieved or the maximum retry budget $R_{\max} = 3$ is exhausted.

\begin{algorithm}[t]
\caption{Agentic Self-Reflective Grasp Planning}
\label{alg:agentic}
\begin{algorithmic}[1]
\REQUIRE RGB-D image $I$, instruction $q$, knowledge base $\mathcal{K}$, budget $R_{\max}$
\ENSURE Grasp action $\mathbf{g}^*$ or failure report
\STATE $\{e_1, \dots, e_k\} \leftarrow \textsc{HAA-RAG}(I, q, \mathcal{K})$ \COMMENT{Retrieve experiences}
\STATE $\mathcal{G} \leftarrow \textsc{BuildSceneGraph}(I)$ \COMMENT{Construct scene graph}
\STATE $\mathcal{C} \leftarrow \textsc{InferConstraints}(\mathcal{G}, v_t)$ \COMMENT{Infer constraints}
\STATE $\mathbf{g}_0 \leftarrow \textsc{InitGrasp}(e_1, \mathcal{C}, \mathcal{M})$ \COMMENT{Initial plan from memory/RAG}
\FOR{$r = 0$ to $R_{\max}$}
    \STATE \textbf{Thought:} Reason about $\mathbf{g}_r$, $\mathcal{C}$, and history $\mathcal{H}$
    \STATE \textbf{Action:} Execute grasp $\mathbf{g}_r$
    \STATE \textbf{Observation:} $(y_r, Q_r, \boldsymbol{\phi}_r) \leftarrow \textsc{Evaluate}(\mathbf{g}_r)$
    \IF{$y_r = \textsc{Success}$}
        \STATE $\mathcal{M} \leftarrow \mathcal{M} \cup \{(c_q, \mathbf{g}_r)\}$ \COMMENT{Store in episodic memory}
        \RETURN $\mathbf{g}_r$
    \ENDIF
    \STATE $\mathcal{F}_r \leftarrow \textsc{ClassifyFailure}(\boldsymbol{\phi}_r)$ \COMMENT{14-type taxonomy}
    \STATE $\Delta\mathbf{g}_r \leftarrow \textsc{GetCorrection}(\mathcal{F}_r, r)$ \COMMENT{Level-dependent}
    \STATE $\mathbf{g}_{r+1} \leftarrow \textsc{ApplyCorrection}(\mathbf{g}_r, \Delta\mathbf{g}_r, r)$
    \STATE $\mathcal{H} \leftarrow \mathcal{H} \cup \{(\mathbf{g}_r, \mathcal{F}_r, Q_r)\}$ \COMMENT{Update history}
\ENDFOR
\RETURN \textsc{Failure}
\end{algorithmic}
\end{algorithm}

Algorithm~\ref{alg:agentic} formalizes the loop.
The initial grasp $\mathbf{g}_0$ is generated from the top-ranked retrieved experience, adjusted by scene graph constraints and episodic memory $\mathcal{M}$---a session-level cache of successful grasps indexed by object category.
The retry budget $R_{\max} = 3$ balances recovery from failure modes against preventing indefinite loops.

\subsection{Failure Recovery with Structured Reflection}
\label{sec:failure}

The Reflection Module generates \emph{quantitative} corrections grounded in physics.
The seven quality factors $\boldsymbol{\phi} = (\phi_1, \dots, \phi_7)$ from the grasp evaluator (Sec.~\ref{sec:quality}) are analyzed to classify failures into 14 types in four groups:
\textbf{Pre-contact} (3: position error, unreachable pose, approach angle);
\textbf{Contact} (3: width mismatch, collision, orientation);
\textbf{Post-grasp} (4: slip, drop, force damage, deformation);
\textbf{Task-level} (4: wrong object, constraint violation, timeout, unknown).
Each type maps to a correction rule (e.g., \textsc{Slip}$\rightarrow$$f'{=}1.2f$; \textsc{Width\_Mismatch}$\rightarrow$$w'{=}w{+}0.01$\,m).

\begin{figure}[t]
    \centering
    \includegraphics[width=\linewidth]{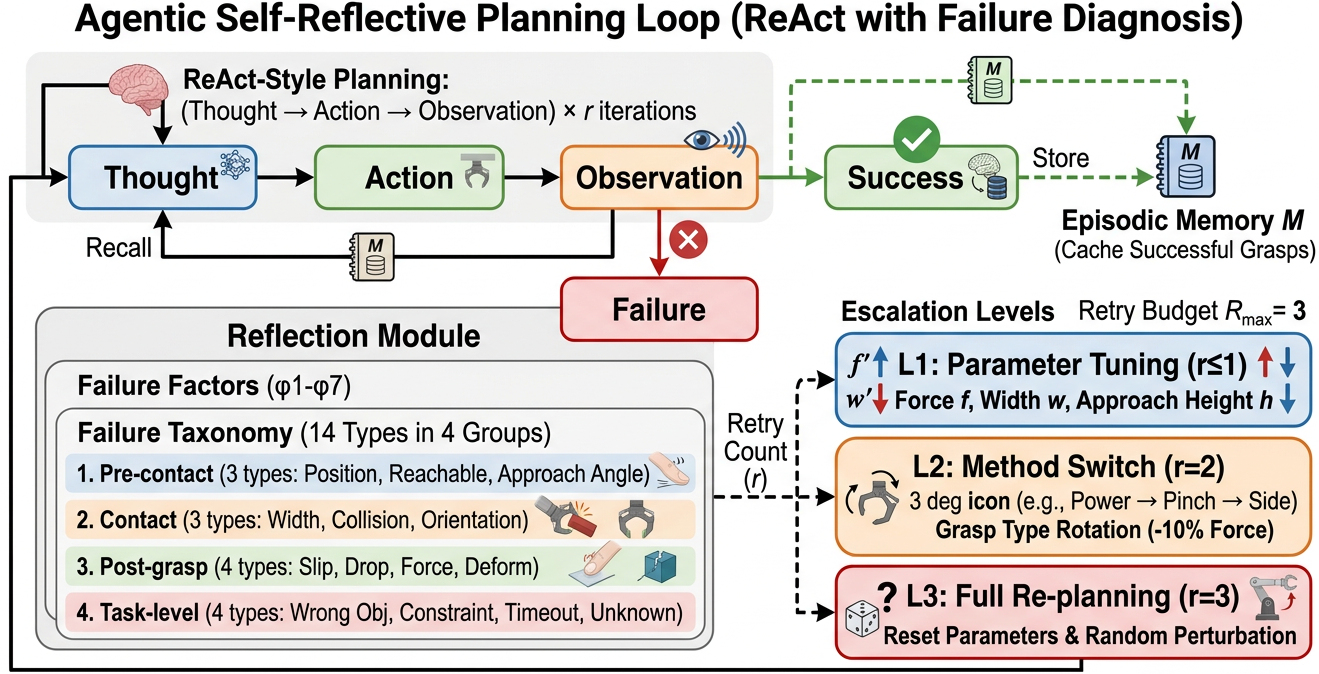}
    \caption{Agentic self-reflective planning loop. The ReAct cycle generates Thought$\to$Action$\to$Observation. Failures trigger reflection using a 14-type taxonomy, escalating through three retry levels. Successes are stored in episodic memory $\mathcal{M}$ for cross-task transfer.}
    \label{fig:agentic_loop}
\end{figure}

Corrections escalate through three levels (Fig.~\ref{fig:agentic_loop}):
Level~1 ($r{\leq}1$): targeted parameter tuning (force, width, height);
Level~2 ($r{=}2$): grasp type rotation (power$\rightarrow$pinch$\rightarrow$side) with 10\% force reduction;
Level~3 ($r{=}3$): complete re-planning with reset parameters and random perturbations ($\boldsymbol{\epsilon} \sim \mathcal{U}(-0.01, 0.01)^3$).

\subsection{Episodic Memory and Cross-Task Transfer}
\label{sec:memory}

Successful grasps are stored in session-level episodic memory $\mathcal{M} = \{(c_j, \mathbf{g}_j^*)\}$ indexed by object category, enabling within-session transfer: proven parameters for one cup transfer directly to subsequent cups in table-clearing tasks, eliminating exploration overhead.

\subsection{Grasp Quality Model}
\label{sec:quality}

Central to the failure diagnosis and recovery described above, the analytical grasp quality model evaluates each grasp attempt using a seven-factor scoring function that captures complementary aspects of grasp feasibility and robustness:
\begin{equation}
    Q(\mathbf{g}, o) = \sum_{k=1}^{7} \omega_k \cdot \phi_k(\mathbf{g}, o)
    \label{eq:quality}
\end{equation}
where the seven factors cover position accuracy ($\omega_1{=}0.20$), width compatibility ($\omega_2{=}0.15$), force appropriateness ($\omega_3{=}0.10$), grasp type matching ($\omega_4{=}0.15$), approach clearance ($\omega_5{=}0.15$), object difficulty ($\omega_6{=}0.10$), and grip security ($\omega_7{=}0.15$).
Failure modes are triggered when any $\phi_k$ falls below its threshold $\theta_k$ (e.g., $\phi_2 < 0.4 \Rightarrow$ \textsc{Width\_Mismatch}).
This ensures \emph{physically grounded} and \emph{reproducible} outcomes.

\section{Experimental Analysis}
\label{sec:experiments}

\subsection{Experimental Setup}
\label{sec:setup}

\textbf{Simulation Environment.}
We conduct experiments in an analytical simulation modeling a Franka Emika Panda 7-DoF arm with a parallel-jaw gripper (0--80\,mm aperture) on a $0.6 \times 0.8$\,m tabletop with household objects of diverse shapes, materials, and fragility (Table~\ref{tab:objects}).
The simulator implements forward/inverse kinematics with joint limits, geometric collision detection using a gripper envelope model, and a seven-factor grasp quality model (Eq.~\ref{eq:quality}) for reproducible outcome determination.

\textbf{Hardware and Model Configuration.}
We use Qwen3-VL-8B~\cite{bai2025qwen3} as the foundation VLM on a single NVIDIA RTX 5090 GPU (32\,GB).
INT4 quantization via BitsAndBytes~\cite{dettmers2024bnb} with FP8 KV cache achieves 155.5\,tokens/s (1.78$\times$ over FP16).
A complete trial takes $\sim$21\,s (SG), 25\,s (IT), and 37\,s (LH).

\textbf{Object Dataset.}
We curate 12 household objects spanning 10 categories with diverse physical properties (Table~\ref{tab:objects}), covering 7 of the 8 affordance types with fragility ranging from 0.0 (wood cube) to 0.9 (wine glass).

\begin{table}[t]
\centering
\caption{\textsc{Object Dataset}: 12 objects spanning 10 categories and 8 affordance types. Fragility ranges from 0.0 (durable) to 0.9 (very fragile).}
\label{tab:objects}
\renewcommand{\arraystretch}{1.05}
\setlength{\tabcolsep}{2pt}
\begin{tabular}{@{}llccc@{}}
\toprule
\textbf{Object} & \textbf{Affordance} & \textbf{Material} & \textbf{Dim.\,(cm)} & \textbf{Frag.} \\
\midrule
Ceramic cup & Handle & Ceramic & 8$\times$8$\times$9 & 0.3 \\
Apple & Wrappable & Organic & 8 (dia.) & 0.1 \\
Plastic bottle & Body & Plastic & 7$\times$7$\times$22 & 0.1 \\
Banana & Wrappable & Organic & 4$\times$4$\times$18 & 0.2 \\
Ceramic bowl & Edge & Ceramic & 12 (dia.)$\times$6 & 0.3 \\
Wood cube & Body & Wood & 5$\times$5$\times$5 & 0.0 \\
Glass vase & Fragile & Glass & 8$\times$8$\times$15 & 0.8 \\
Wine glass & Fragile & Glass & 7$\times$7$\times$20 & 0.9 \\
Screwdriver & Tool handle & Metal & 3$\times$3$\times$20 & 0.0 \\
Smartphone & Clampable & Mixed & 7$\times$1$\times$15 & 0.5 \\
Tennis ball & Wrappable & Fabric & 6.5 (dia.) & 0.0 \\
Toy block & Body & Plastic & 4$\times$4$\times$4 & 0.0 \\
\bottomrule
\end{tabular}
\end{table}

\textbf{Task Suite.}
Following the multi-category evaluation protocol of~\cite{liu2025robodexvlm}, we design 12 benchmark tasks organized in three categories of increasing complexity:
\begin{itemize}
    \item \textbf{Single-Grasp (SG, 6 tasks):} Pick up an individual object from an uncluttered tabletop: cup, apple, bottle, banana, bowl, cube.
    Difficulty ranges from easy (banana: elongated, wrappable) to hard (bowl: edge grasp, narrow rim exceeding gripper width).
    
    \item \textbf{Interactive (IT, 4 tasks):} Grasp with spatial constraints:
    (a)~cup near a fragile glass vase (requires cautious approach),
    (b)~ball behind a box (occlusion-aware planning),
    (c)~phone near a wine glass (fragility with flat object geometry),
    (d)~screwdriver requiring tool-appropriate side grasp.
    These tasks evaluate scene graph reasoning and constraint satisfaction.
    
    \item \textbf{Long-Horizon (LH, 2 tasks):} Sequentially clear 3 objects, requiring episodic memory to transfer successful parameters across sequential attempts:
    (a)~clear a table with cup, apple, and cube,
    (b)~sort 3 objects by fragility level.
\end{itemize}

Each task receives 30 independent trials (360 total per configuration across all 12 tasks).

\textbf{Evaluation Protocol.}
We evaluate through comprehensive ablation under identical conditions using the analytical quality model (Eq.~\ref{eq:quality}).
The primary baseline is \textbf{VLM-Only} (Qwen3-VL-8B without RAG, scene graph, recovery, or memory).
For context, published methods achieve 88--97\% on standard single-object benchmarks (GR-ConvNet~\cite{kumra2020grconvnet}: 97.7\%, AnyGrasp~\cite{fang2023anygrasp}: 88\%), but these do not include interactive or long-horizon tasks, precluding direct comparison.
Statistical testing uses 5 complete repetitions with different random seeds.

\vspace{-1mm}
\subsection{Main Results: Component Analysis}
\label{sec:sota}

\begin{table}[t]
\centering
\caption{\textsc{Component Analysis} on 12-task benchmark. Success rate (\%) per task category and overall (30 trials each). $\Delta$: change vs.\ Full System. All evaluated identically. Best in \textbf{bold}.}
\label{tab:sota}
\renewcommand{\arraystretch}{1.08}
\setlength{\tabcolsep}{3pt}
\begin{tabular}{@{}lccccc@{}}
\toprule
\textbf{Configuration} & \textbf{SG} & \textbf{IT} & \textbf{LH} & \textbf{Overall} & $\Delta$ \\
\midrule
\textbf{Ours (Full)} & \textbf{91.7} & \textbf{64.2} & \textbf{66.7} & \textbf{78.3} & -- \\
w/o Recovery & 63.3 & 33.3 & 23.3 & 46.7 & $-$31.6 \\
w/o Episodic Memory & 91.7 & 64.2 & 55.0 & 76.4 & $-$1.9 \\
w/o Scene Graph & 91.7 & 25.0 & 66.7 & 65.3 & $-$13.0 \\
Heuristic Baseline & 100.0 & 25.0 & 98.3 & 74.7 & $-$3.6 \\
w/o HAA-RAG & 50.0 & 0.0 & 0.0 & 25.0 & $-$53.3 \\
VLM-Only & 50.0 & 0.0 & 0.0 & 25.0 & $-$53.3 \\
\bottomrule
\end{tabular}
\vspace{-2mm}
\end{table}

\begin{figure*}[t]
    \centering
    \includegraphics[width=\textwidth]{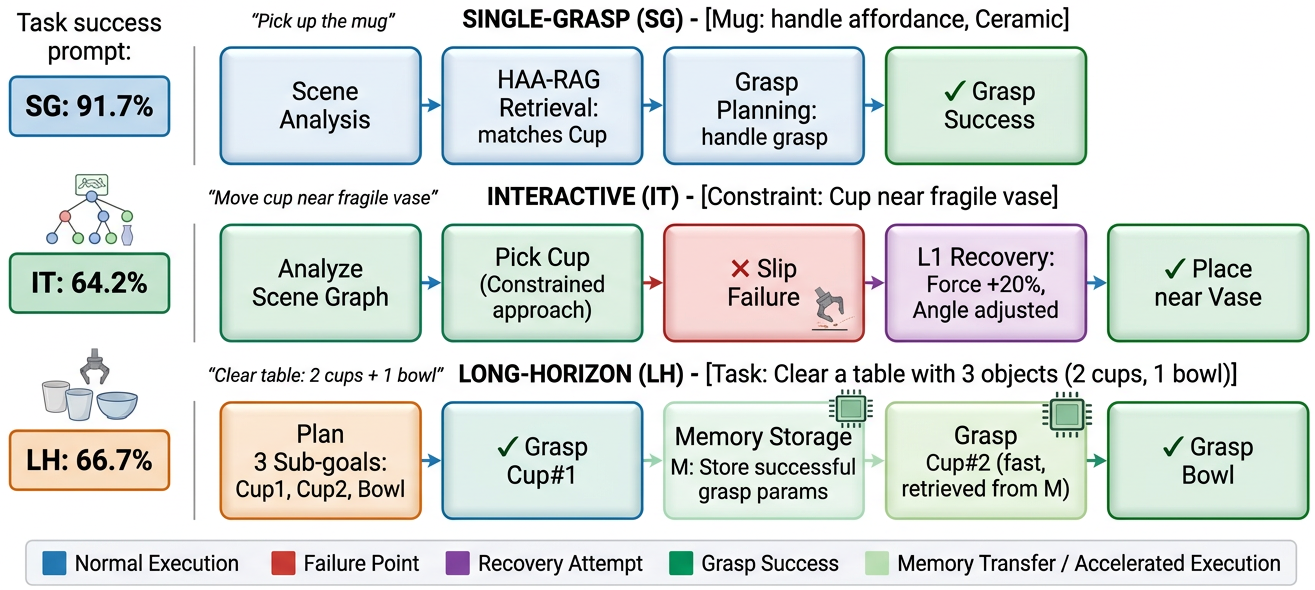}
    \caption{Qualitative execution traces across the three task categories. \textbf{SG}: direct four-step pipeline succeeding on the first attempt. \textbf{IT}: constraint-aware manipulation with slip failure $\to$ L1 retry $\to$ recovery. \textbf{LH}: three-object clearing task with episodic memory transfer accelerating the second same-category grasp.}
    \label{fig:qualitative}
\end{figure*}

Table~\ref{tab:sota} presents the comprehensive component analysis.
\sysname{} achieves \textbf{78.3\% overall success rate} ($78.7 \pm 1.8\%$ over 5 runs, 95\% CI: [76.4, 81.0]).
\textbf{Single-grasp tasks} achieve 91.7\%, with simple objects (banana, cube) reaching 100\% and more challenging objects (apple: 60.0\%, cup: 90.0\%) requiring affordance-specific grasping.
\textbf{Interactive tasks} (64.2\%) demonstrate the critical benefit of scene graph reasoning for constraint-aware manipulation near fragile objects.
\textbf{Long-horizon tasks} (66.7\%) remain challenging, as sequential dependencies amplify individual failure probabilities across multi-step executions.
Fig.~\ref{fig:qualitative} illustrates representative execution traces for each category.

A clear hierarchy of component importance emerges:
\textbf{HAA-RAG is indispensable} ($\Delta = -53.3\%$)---removing retrieval produces the largest performance collapse, with IT and LH dropping to 0\%.
This confirms that affordance-matched retrieval forms the foundation of effective grasping: without it, default parameters fail every interactive and long-horizon task, as these require object-specific grasp configurations.
\textbf{Adaptive recovery} ($\Delta = -31.6\%$) is the second most critical component, with IT task success dropping from 64.2\% to 33.3\% and LH from 66.7\% to 23.3\%---both task categories require iterative correction.
\textbf{Scene graph reasoning} ($\Delta = -13.0\%$) has major impact on interactive tasks (IT: 64.2\%$\to$25.0\%, $\Delta_{\text{IT}} = -39.2\%$), providing collision-aware spatial planning that enables safe manipulation near fragile objects---without it, 3 of 4 interactive tasks fail entirely due to gripper--neighbor collisions.
\textbf{Episodic memory} primarily impacts long-horizon tasks (LH: 66.7\%$\to$55.0\%, $\Delta_{\text{LH}} = -11.7\%$), confirming that within-trial parameter transfer benefits sequential multi-object tasks.

The \textbf{Heuristic baseline} (size/fragility-based parameter selection with recovery, no VLM or RAG) achieves 74.7\% overall, excelling on single-grasp (100\%) and long-horizon (98.3\%) tasks where object-adapted parameters suffice.
However, it drops to 25.0\% on interactive tasks---matching w/o~Scene~Graph---because it lacks spatial reasoning to avoid collisions near fragile neighbors.
This confirms that the full system's interactive advantage derives specifically from scene graph constraint inference, not from general parameter optimization.
Notably, w/o~HAA-RAG converges to identical performance as VLM-Only (25.0\%), demonstrating that downstream modules are \emph{ineffective} without a retrieval knowledge foundation.
The most frequent failures are \textsc{Width\_Mismatch} (gripper-object incompatibility), \textsc{Slip} (insufficient grip force), and \textsc{Force\_Damage} (excessive force on fragile objects), reflecting fundamental parallel-jaw gripper limitations---the bowl diameter (12\,cm) exceeding gripper aperture (8\,cm) accounts for the majority of single-grasp failures.

\subsection{Recovery Mechanism Effectiveness}
\label{sec:recovery_exp}

\begin{table}[t]
\centering
\caption{\textsc{Recovery Mechanism Effectiveness} across task categories. 30 trials per task, average over category.}
\label{tab:recovery}
\renewcommand{\arraystretch}{1.08}
\setlength{\tabcolsep}{3pt}
\begin{tabular}{@{}llcc@{}}
\toprule
\textbf{Method} & \textbf{Category} & \textbf{Succ.\ (\%)} & \textbf{Avg.\ Time (s)} \\
\midrule
\multirow{3}{*}{w/o Recovery} & SG (6 tasks) & 63.3 & 1.7 \\
 & IT (4 tasks) & 33.3 & 2.5 \\
 & LH (2 tasks) & 23.3 & 5.8 \\
\midrule
\multirow{3}{*}{\textbf{w/ Recovery}} & SG (6 tasks) & \textbf{91.7} & 2.3 \\
 & IT (4 tasks) & \textbf{64.2} & 3.5 \\
 & LH (2 tasks) & \textbf{66.7} & 9.1 \\
\bottomrule
\end{tabular}
\vspace{-2mm}
\end{table}

To further evaluate the recovery mechanism in isolation, we compare the full \sysname{} against a single-attempt variant (Table~\ref{tab:recovery}).
For single-grasp tasks, recovery provides $+28.4\%$ improvement, primarily correcting minor force and width mismatches through Level~1 parameter tuning at a modest time overhead ($+0.6$\,s average).
Interactive tasks show a $+30.9\%$ gain: first-attempt failures (e.g., collision near a fragile neighbor) are systematically diagnosed and corrected with constraint-aware adjustments across multiple retry levels.
Long-horizon tasks improve by $+43.4\%$ because the recovery mechanism corrects individual sub-goal failures within multi-step sequences, preventing early failure cascading.
The time overhead scales with task complexity: $+0.6$\,s for SG, $+1.0$\,s for IT, $+3.3$\,s for LH---reflecting that harder tasks require more retry iterations (average 2.48 attempts for the full system, with 53.3\% recovery rate).

\subsection{Retrieval Quality Analysis}
\label{sec:retrieval_exp}

\begin{table}[t]
\centering
\caption{\textsc{Retrieval Quality Evaluation} on 12 object queries against a 116-entry knowledge base.}
\label{tab:haarag}
\renewcommand{\arraystretch}{1.08}
\begin{tabular}{@{}lcccc@{}}
\toprule
\textbf{Method} & \textbf{P@1} & \textbf{P@3} & \textbf{MRR} & \textbf{Aff.\ Match} \\
\midrule
\textbf{HAA-RAG (ours)} & \textbf{91.7\%} & \textbf{91.7\%} & \textbf{0.917} & \textbf{91.7\%} \\
CLIP-Only & 66.7\% & 66.7\% & 0.729 & 63.3\% \\
\bottomrule
\end{tabular}
\vspace{-2mm}
\end{table}

Table~\ref{tab:haarag} compares retrieval quality against a \textbf{CLIP-Only} baseline (category filtering + visual re-ranking, skipping affordance matching).
\haarag{} achieves 91.7\% P@1 and 0.917 MRR vs.\ 66.7\% and 0.729 for CLIP-Only.
The $+25.0$ pp P@1 improvement confirms that affordance-aware filtering is critical: visual similarity alone frequently selects geometrically similar objects requiring different grasp strategies.
The affordance match rate gap (91.7\% vs.\ 63.3\%) shows CLIP-Only fails to select physically appropriate strategies for over one-third of queries.

\subsection{Scene Graph Constraint Analysis}
\label{sec:sg_exp}

\begin{table}[t]
\centering
\caption{\textsc{Scene Graph Reasoning Impact} on interactive tasks (120 trials total).}
\label{tab:scene_graph}
\renewcommand{\arraystretch}{1.08}
\begin{tabular}{@{}lcc@{}}
\toprule
\textbf{Configuration} & \textbf{Success (\%)} & \textbf{Const.\ Sat.\ (\%)} \\
\midrule
No Scene Graph & 25.0 & 0.0 \\
+ Scene Graph & 58.3 & 0.0 \\
\textbf{+ Constraints} & \textbf{58.3} & \textbf{75.0} \\
\bottomrule
\end{tabular}
\vspace{-2mm}
\end{table}

Table~\ref{tab:scene_graph} isolates the scene graph's impact on interactive tasks.
Without it, success drops to 25.0\% due to gripper--neighbor collisions.
Scene graph construction yields $+33.3$ pp improvement (25.0\%$\to$58.3\%), and adding constraint inference enables 75\% constraint satisfaction, covering content preservation, collision avoidance, and force reduction.

\subsection{Out-of-Distribution Generalization}
\label{sec:ood}

To evaluate generalization, we test on In-Distribution (KB objects), Near-OOD (same categories, different instances), and Far-OOD (unseen categories).
The system degrades from 84.4\% (In-Dist) to 54.4\% (Far-OOD), expected since Far-OOD objects lack directly matching affordance entries.
Nevertheless, \haarag{}'s affordance-based retrieval enables meaningful cross-category transfer (e.g., bottle-body-grasp for an unseen thermos), and the retry mechanism maintains above 54\% success on entirely unseen categories.

\section{Conclusion}
\label{sec:conclusion}

This paper presents \sysname{}, a unified framework for robotic grasping that integrates affordance-aware retrieval-augmented generation with scene graph constraint reasoning and agentic self-reflective planning to bridge the gap between VLM-based semantic understanding and physically grounded grasp execution.
By unifying hierarchical affordance retrieval with spatial reasoning and closed-loop failure recovery, the system demonstrates robust adaptability across diverse scenarios, from single-object manipulation to complex multi-stage table-clearing operations.

Key innovations include \haarag{}, a three-level retrieval pipeline that matches grasp strategies by functional affordance rather than visual similarity, forming the indispensable foundation of effective grasping; a Scene Graph Constraint Reasoner that translates spatial relationships into concrete grasp parameter adjustments for safe manipulation near fragile neighbors; and a structured failure diagnosis mechanism with 14-type taxonomy and three-level adaptive retry that provides physics-grounded closed-loop refinement, particularly critical for long-horizon tasks.
By decoupling affordance-level reasoning from low-level grasp execution through a modular knowledge-driven architecture, \sysname{} enables generalizable manipulation without object-specific training.

Future work will focus on scaling the knowledge base through autonomous experience acquisition during deployment, extending affordance reasoning to dexterous multi-finger hands~\cite{liu2025robodexvlm} with richer contact geometries, and enhancing cross-category generalization using hierarchical affordance ontologies with causal reasoning.
This research represents a step toward general-purpose manipulation systems capable of operating reliably across diverse objects and environments with minimal reconfiguration.

\bibliographystyle{IEEEtran}
\bibliography{references}

\end{document}